\documentclass[11pt,a4paper]{article}
\usepackage[hyperref]{emnlp2020}
\usepackage{times}
\usepackage{hhline}
\usepackage{listings}
\usepackage{latexsym}
\usepackage{graphicx}

\usepackage{microtype}

\usepackage{comment}
\usepackage{cleveref}
\usepackage{tikz}
\usepackage{subcaption}
\usepackage{fancyvrb}

\aclfinalcopy 

\title{Transferring Knowledge from Vision to Language:\\How to Achieve it and how to Measure it?}

\author{Tobias Norlund\thanks{\hspace{0.14cm}Equal contribution.}\hspace{0.05cm} \textsuperscript{\dag} \hspace{4mm} Lovisa Hagström\footnotemark[1] \hspace{4mm} Richard Johansson \\
  Division of Data Science and Artificial Intelligence \\
  Chalmers University of Technology \\
  \textsuperscript{\dag} Recorded Future \\
  \texttt{\{tobiasno, lovhag, richajo\}@chalmers.se}}

\date{}

\begin{document}
\maketitle
\begin{abstract}
Large language models are known to suffer from the hallucination problem in that they are prone to output statements that are false or inconsistent, indicating a lack of knowledge. A proposed solution to this is to provide the model with additional data modalities that complements the knowledge obtained through text. We investigate the use of visual data to complement the knowledge of large language models by proposing a method for evaluating visual knowledge transfer to text for uni- or multimodal language models. The method is based on two steps, 1) a novel task querying for knowledge of memory colors, i.e. typical colors of well-known objects, and 2) filtering of model training data to clearly separate knowledge contributions. Additionally, we introduce a model architecture that involves a visual imagination step and evaluate it with our proposed method. We find that our method can successfully be used to measure visual knowledge transfer capabilities in models and that our novel model architecture shows promising results for leveraging multimodal knowledge in a unimodal setting.

\end{abstract}

\section{Introduction}
Large language models have proved performant across a diverse set of tasks in NLP, and most recently even as unsupervised multitask learners~\citep{radford2019language, gpt3}.
An important contributing factor to this is the capability of the models to hold large amounts of linguistic as well as factual knowledge in their parameters.

While impressive, without strong task-specific fine-tuning these models are prone to outputting false or inconsistent statements, often also referred to as \emph{hallucination}~\citep{logan-etal-2019-baracks}.
This has been particularly studied for generative tasks such as abstractive text summarization~\citep{maynez-etal-2020-faithfulness} and dialog systems~\citep{roller-etal-2021-recipes, li-etal-2020-dont}, but the problem is also apparent for models applied to cloze-style fill-in-the-blank tasks~\citep{petroni-etal-2019-language, jiang-etal-2020-know}.
Having truthful NLP systems is a core requirement for most applications, which is why this is an important problem to address.

Grounding has been proposed as a potential way to mitigate this problem, e.g by providing broader world information from for example multimodal perception~\citep{bisk-etal-2020-experience, bender-koller-2020-climbing}.

Information from multimodal perception may actually provide a significant amount of additional world information to an NLP model, since text data suffers from the problem of \emph{reporting bias}. That is, humans generally communicate novel information rather than trivial, leading to a discrepancy between reality and what gets described in text~\citep{gordon2013reporting}. Consequently, perceptual information may contain complementing world knowledge that cannot be found in text data, and has the potential to mitigate the aforementioned problem of hallucinating NLP models.

Previous works have evaluated how grounded language representations impact performance on common NLP benchmarks~\citep{sileo2021visual, kiela-etal-2018-learning, elliott-kadar-2017-imagination}, but little has been done on investigating grounding specifically as an additional source of knowledge.

In this work, we take a focused look at how data from a visual modality can augment the knowledge a language model expresses.
We design an experimental setup to enable the development of strategies for maximizing visual-to-textual knowledge transfer.
In the setup, we create a small knowledge-centric cloze-style task in English named \emph{Memory Colors} that is tailored to test for visual knowledge by querying for the typical colors of well-known items.
We also build a large vision-and-language dataset in the English language, where we carefully control for the modality from which the necessary visual knowledge can be learnt.
Finally, we use this data to train self-supervised multimodal models, and  compare strategies to query for the visual knowledge.

Based on intuitions of how humans are able to store and retrieve such knowledge, we also propose a querying strategy that involves ``imagining'' a visual representation from which the answer then can be decoded.

To summarize, our contribution is twofold:
\begin{enumerate}\addtolength{\itemsep}{-0.5\baselineskip}
    \item We provide an experimental setup for evaluating visual knowledge transfer in English multimodal language models, including a novel task we denote \emph{Memory Colors}.
    \item We propose a language model querying strategy involving a visual imagination step and show that it can provide an efficient means of visual knowledge transfer compared to standard querying.
\end{enumerate}

\section{An experimental setup evaluating visual knowledge transfer} \label{sec:eval_method}

Humans have the ability to learn knowledge from non-linguistic modalities (such as visual perception) and express this in language, making them able to e.g. textually reason about what an elephant looks like because they have previously seen said animal in an image. 
Many models that integrate the textual and visual modalities exist, but the majority of them have been created with the purpose of reasoning about 
properties of \emph{individual images} provided to the system: for instance, to ask about an elephant, you need to simultaneously provide the model with an image of an elephant. 

We hypothesize that the capability to incorporate knowledge from different modalities and expressing it textually could improve on the common sense as well as in-domain knowledge that language models possess.
To this end, we wish to create an experimental setup in which we can measure how well a model can acquire visual knowledge and then express it in text.

A simple way to evaluate a model for its capability to transfer visual knowledge into text is to query it about typical colors of certain objects -- \emph{memory colors} -- while making sure that the model cannot acquire this knowledge through a text signal, i.e. it has not previously been told what the colors should be. 

Consequently, we create a zero-shot cloze-style task of predicting memory colors of common objects, described in \Cref{sec:memory_colors}. We also collect a large vision-and-language dataset for model training in which we carefully control for whether the knowledge necessary for solving the memory color task is available strictly in the visual, textual, or both modalities, described in \Cref{sec:vision-language-pre-training-dataset}.

\subsection{Memory Colors dataset}\label{sec:memory_colors}

When human observers can agree on the typical color, or canonical color, of a certain object type through their experiences with instances of said object type, the color of that object is generally referred to as its \emph{memory color}~\citep{perez1998familiar}. For example, a banana can be green or brown, but it is usually remembered as being yellow, such that \emph{yellow} is the memory color of a banana. As explained by \citet{newhall1957comparison} \mbox{``... color memory} is a selective resultant of the relative
impressiveness during perception of the various aspects of stimulation. More dominant,
characteristic, and attractive aspects tend to be more impressive, and less dominant aspects
tend to be less impressive. The more impressive aspects are more prone to survival in
subsequent memory while other aspects are not.''

As such, memory colors of typical objects are remembered by humans and a human can answer questions about what the typical color of such an object is despite not having the object in front of them when answering. Consequently, memory colors express visual knowledge and we can use them for a simple zero-shot evaluation of whether a model can display the same capability as a human of transferring a visual signal into memory and, later on, text.

\begin{figure}[h]
    \centering
    \vspace{-5mm}    
    \input{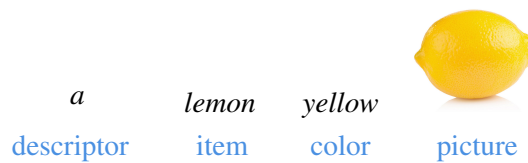}
    \caption{One entry in the Memory Colors dataset.}
    \label{fig:data-example}
\end{figure}

For our visual knowledge transfer evaluation task we create a novel Memory Colors dataset in the English language, consisting of 109 object types paired with their memory color, an illustrating picture and a descriptor. 

\Cref{fig:data-example} shows an example, and the supplementary material includes the full dataset with additional statistics.

The Memory Colors dataset and a corresponding human baseline is obtained by annotating a set of randomly shuffled cloze questions based on well-known entities with typical colors. Examples of such entities are items, materials, animals, ingredients or plants that are observable in the real world, such as \emph{tomato}, \emph{elephant}, \emph{cocoa} and \emph{grass}. These entities were sourced from the web, including Wikidata\footnote{\url{www.wikidata.org}} and ConceptNet,\footnote{\url{www.conceptnet.io}} as well as from the commonsense knowledge of the authors of this article.

The cloze questions of the Memory Colors dataset are created with the help of a predefined query template; see an example question in \Cref{tab:example-question}. The predefined query template is assigned to each annotator from a set of seven different templates to create differently formatted questions querying for the same visual knowledge.

The memory colors used for the items in the dataset are black, blue, brown, green, grey, orange, pink, purple, red, white and yellow. The annotators are asked to pick their answer from one of these 11 colors for each question. They are also asked to answer the questions to the best of their ability, without consulting other information sources.\footnote{All the query templates and annotator instructions are provided in the supplementary material.}

\paragraph{Memory color label} The color label for each item is given by the majority vote of 11 annotators, and only items with a minimum of 8 annotators agreeing on a memory color are included in the dataset, resulting in the Memory Colors dataset consisting of 109 items and corresponding memory colors, with a majority vote distribution as indicated in \Cref{tab:vote-count}. 

\begin{table}[h]
    \centering
    \caption{The count of the number of majority votes for each of the items in the Memory Colors dataset. A majority vote of 11 means that all annotators agreed on the color.}
    \begin{tabular}{c|c}
    \hline
    \# of majority votes & Count \\
    \hline
        11 & 60 \\
        10 & 32 \\
        9 & 7 \\
        8 & 10 \\
    \hline     
    \end{tabular}
    \label{tab:vote-count}
\end{table}

Arguably, our use of the term \emph{memory color} may be somewhat less strict than that of the optical science field, in which very few memory colors are admitted due to high requirements on agreement between humans for a color to be classified as a memory color. However, for the sake of obtaining a dataset of a sufficent size, we decide to also include items and colors for which there is a majority, while not a perfect one.

\begin{table}[h]
    \centering
    \caption{An example of a cloze-question provided to a human annotator, given by the query template \emph{Q: What is the color of [DESCRIPTOR] [ITEM]? A: [MASK].}}
    \begin{tabular}{l|c}
    \hline
    Question & Answer \\
    \hline
    Q: What is the color of \emph{a} \emph{lemon}? &  \emph{yellow} \\ A: [answer] & \\
    \hline
    \end{tabular}
    \label{tab:example-question}
\end{table}

\paragraph{Descriptor} The descriptor for each item in the dataset is manually added to make the cloze-questions grammatically correct and to resolve potential item reference ambiguities. For example, determiners such as ``a'', ``an'' or ``
the'' are added as a descriptor for countable nouns and the addition ``the animal'' might be added for the occurence \emph{seal} to clarify that we refer to the animal and not e.g. a letter seal.

\paragraph{Illustrating picture} A picture of each item in the dataset is manually added by the authors by picking an image from the Internet that is deemed to correspond well to the item, and that to the authors' best ability reflects the labeled color.

\paragraph{Human baseline} The human baseline for the task is taken as the mean of the accuracy scores of 11 annotators, where each accuracy score is calculated by comparing the annotator answers with the majority vote labels. The annotators achieved a mean accuracy score of 0.937 with a standard deviation of 0.051 on the Memory Colors dataset. 

We hypothesize that a perfect accuracy score is not reached due to different perceptions of colors, varying knowledge of what the Memory Colors items refer to and that it perhaps is unavoidable that some disagreements exist for this fairly large dataset that has an apparent dependence on cultural background. 

Hypothetically, the phrasing of the question may also be a factor that explains some of the variation \cite{kalton1982effect}, although this seems unlikely in this case since the questions concern concrete physical objects.

\paragraph{Annotator agreement} To verify our Memory Colors dataset and its human baseline we also evaluate the annotator agreement between the 11 annotators using Fleiss' kappa score~\citep{fleiss1971measuring}. The kappa score for the agreement between the annotators is found to be 0.863, indicating that the annotators agree fairly well.

\subsection{Vision-and-language dataset}\label{sec:vision-language-pre-training-dataset}
\begin{table}[t]
\caption{\label{tab:vlp_dataset_stats_small} Total number of image and caption samples for the full and filtered versions of the training dataset.}
\begin{center}
\begin{tabular}{l|l}
\hline
                        & \textbf{Total}      \\ \hhline{=|=}
\textbf{Validation}     &                     \\
\hspace{2mm} Captions   & 58,937              \\
\hspace{2mm} Images     & 38,234              \\

\textbf{Training-unfiltered}    &                     \\
\hspace{2mm} Captions   & 4,720,971           \\
\hspace{2mm} Images     & 2,911,438           \\

\textbf{Training-filtered} &                     \\
\hspace{2mm} Captions  & 4,429,671           \\
\hspace{2mm} Images     & 2,749,612           \\  \hline
\end{tabular}
\end{center}
\end{table}
We combine four public image+text datasets to be used for self-supervised training in our experiments: MS COCO~\citep{coco}, SBU Captions~\citep{sbu_captions}, Visual Genome QA~\citep{krishnavisualgenome} and Conceptual Captions~\citep{sharma2018conceptual}.
In total it comprises 4.7M captions paired with 2.9M unique images.

At the core of this work is a method to measure visual knowledge transfer by means of the memory colors task described in Section \ref{sec:memory_colors}.
To this end, we construct a version of this vision-and-language dataset where we 
remove training examples in which a memory color is revealed in the caption.
This way we can, with high confidence, attribute correct model predictions to originate from the visual modality rather than the captions.
In the filtered version, an example is excluded if its caption matches either of two conditions:
\begin{enumerate}\addtolength{\itemsep}{-0.5\baselineskip}
    \item It contains any object word  and any color word from the memory colors dataset, by exact string match. 
    \item When tokenized and stemmed, it contains any stemmed object word and any stemmed color word from the memory colors dataset
\end{enumerate}
The above filter matches about 6\% of the captions in the training set.
A summary of the statistics for the full and filtered versions of the training dataset are detailed in \Cref{tab:vlp_dataset_stats_small}. Complete statistics for the dataset can be found in the supplementary material.

\section{Models}

\begin{figure}[t]
    \includegraphics[width=77mm]{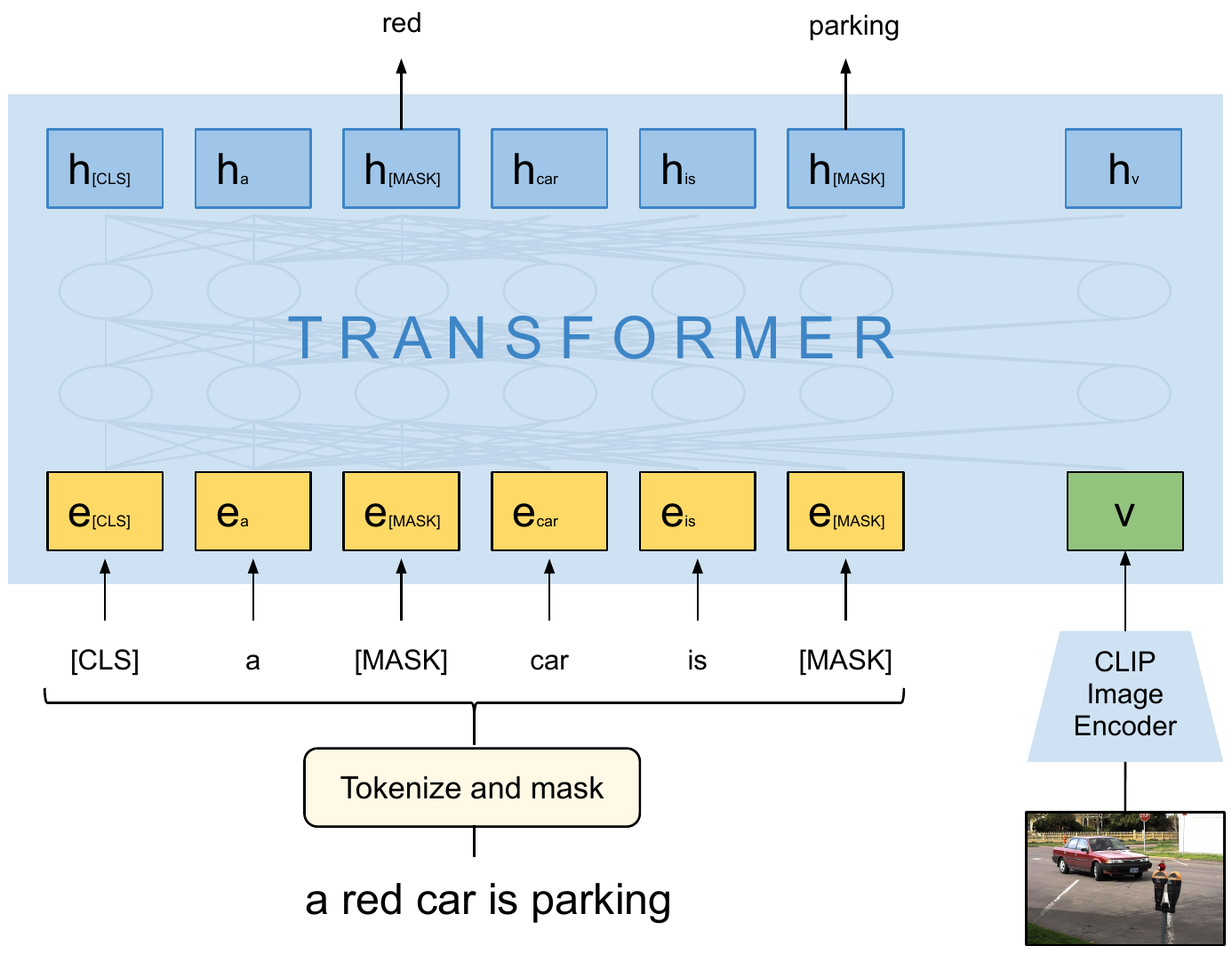}
    \caption{Training of the multimodal CLIP-BERT model using MLM. An image represented by CLIP is appended to the transformer input.}
    \label{fig:clip_bert_pretraining}
\end{figure}
\begin{figure*}
    \includegraphics[width=16cm]{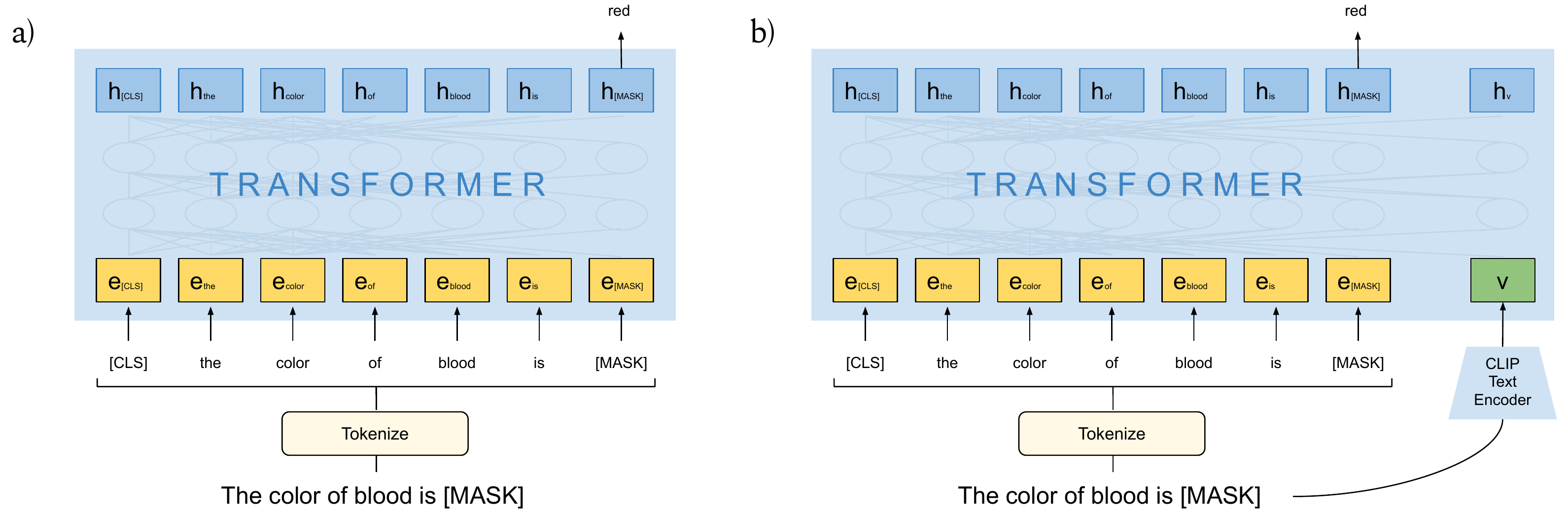}
    \caption{a) Inference from CLIP-BERT using the implicit transfer strategy, directly querying for knowledge through a masked token prediction. b) Inference from CLIP-BERT using the explicit transfer strategy, involving the prediction of visual latent features (visual imagination) as a preceding step.}
\end{figure*}
When a human is asked a question like \emph{What is the color of your house?} it typically requires retrieving a mental picture from memory of what the house looks like. 
Based on this mental picture, the answer can then easily be inferred.
The mental picture provides an efficient means to store knowledge about the appearance of the house, as other questions like \emph{How many floors does it have?} or \emph{Does it have a garden?} can as easily be inferred.
We hypothesize that this idea of visual imagination could also provide an efficient means of visual knowledge transfer, and propose a model for performing the ``imagination'' explicitly.

We take inspiration from recent works in vision-and-language modeling, where the transformer architecture~\citep{attention_is_all} has become the de facto standard ~\citep{NEURIPS2019_c74d97b0,tan-bansal-2019-lxmert,qi2020imagebert,li2020oscar}.
In a typical setup, an image representation is fed to the transformer encoder jointly with the text tokens, and the encoder is then pre-trained
using various denoising and contrastive objectives.

In this work, we perform experiments on a simple yet novel variant to accomodate for the visual imagination.
While common practice is to use visual features from an object detector~\citep{ren2015faster}, we extract visual representations using the image encoder of a pretrained CLIP model~\citep{radford2021learning} instead.
CLIP consists of two networks for encoding an image and a text sentence respectively, and is trained to align these representations in a joint space using a contrastive training objective.
The resulting visual encoder is shown to have great discriminatory performance, for example when applied to zero-shot image classification. However, the main benefit of using CLIP to extract visual features is the joint feature space between its visual and textual encoder, enabling us to generate ``visual'' features from text.

In our experiments, we start from the popular pretrained BERT base model\footnote{\texttt{bert-base-uncased} in Huggingface Transformers.}, and continue training on our visual-language dataset from \Cref{sec:vision-language-pre-training-dataset}, using only the Masked Language Modelling (MLM) objective with 15\% random dynamic masking ratio.
Specifically, we train two models on the filtered and unfiltered versions respectively:

\paragraph{BERT-base} We continue training of BERT base using MLM \emph{only on the captions part} of the visual-and-language dataset. 
This provides a baseline of the amount of color knowledge that can be picked up from text alone.
                             
\paragraph{CLIP-BERT} We continue training of BERT base using MLM but \emph{on both the captions and the images} of the visual-and-language dataset. 
The image representation is transformed through a projection layer and appended to the transformer input without adding any positional or segment embeddings.
The MLM objective is only applied on the textual positions.
An illustration of the training of CLIP-BERT is shown in \Cref{fig:clip_bert_pretraining}.

\subsection{Training}
All models were trained for 16 hours using 32 16 GB T4 GPUs with a total batch size of 16,384.
During this time between 44k to 58k gradient steps were taken, and all validation losses had converged.
We used the Adam optimizer with a constant learning rate of 5e-5, and applied mixed-precision training for increased performance.

\subsection{Querying strategies}\label{sec:querying-strategies}
The canonical way to query BERT-like models for knowledge in a zero-shot setting is to construct textual templates containing a \texttt{[MASK]} token to be predicted by the model in a cloze-style fashion~\citep{petroni-etal-2019-language}.
Similarly, we manually construct templates to query for the color of objects in Memory Colors.
Since it has been shown that language models can be sensitive to the exact phrasing of such templates~\citep{jiang-etal-2020-know}, we construct a set of 13 distinct alternatives paraphrasing each other. The templates provided to the human annotators (described in \Cref{sec:memory_colors}) are included in these alternatives, while the model templates are complemented with versions that also contain model-specific tokens, such as \texttt{{[}SEP{]}}. All model templates are listed in the supplementary material.
We report the mean top-1 accuracy and standard deviation of each model over all templates, and we only consider the eleven valid color words from the full vocabulary of model predictions.

Since the goal of our work is to investigate how visual knowledge can be transferred into language models, we consider two mechanisms of knowledge transfer, denoted \emph{implicit} and \emph{explicit} transfer respectively. These mechanisms are investigated using two different querying strategies.

\subsubsection{Implicit transfer}

By \emph{implicit transfer}, we refer to the effect of multimodal training on the word representations of a language model.
To measure the implicit transfer capabilities of a model, we use a multimodal signal at training time but at test time, we query the model as described above using the method proposed by \citet{petroni-etal-2019-language}.
We use the term \emph{implicit}, as the visual knowledge (e.g. that blood typically has a red hue) needs to be memorized in the model weights as a part of MLM training, and later retieved textually (the correct masked token should be ``red'').

\subsubsection{Explicit transfer by visual imagination}
As an alternative to implicit transfer, we propose a more explicit transfer strategy where we as a preceeding step predict visual features of an imaginary image based on the text.\footnote{We refer to the raw text, including the [MASK] part.}
These predicted features are then appended to the transformer input that thus becomes complete with both the textual and visual features as seen during training.
For this visual prediction, we use the textual encoder of CLIP as it is explicitly trained to align its representations with the visual counterpart.

To evaluate the quality of the predicted representations on the Memory Colors dataset, we also generate ``true'' visual representations with the visual encoder of CLIP using a ground truth image of each object, and evaluate each model using these as well.
This setting more resembles visual question answering, and should be considered as an upper bound for what performance can be expected from the predicted features.

\section{Results and analysis}
We evaluate the transfer capabilities of our aforementioned models both to assess the functionality of our measurement method and to investigate the effect of implicit and explicit visual knowledge transfer. 
The results on Memory Colors for the different experiments are displayed in \Cref{tab:memory-colors-results}. 
We structure the analysis in this section around a set of interesting questions.

\begin{table}[h]
    \centering
    \caption{The mean and standard deviation of the accuracy scores of the models on the Memory Colors dataset for different query templates. 
    CLIP-BERT-images is the only model that is given the pictures from the dataset during evaluation.}
    
    \begin{tabular}{l l | c c}
    \hline
    Training & Model & Accuracy \\
    \hhline{==|=}
     & Random baseline & $0.091 \pm 0.026$ \\
     & Majority baseline & $0.229\pm0.000$ \\
     & Human baseline    & $0.937 \pm 0.051$ \\
    \hline
    None & BERT-base & $0.252 \pm 0.102$ \\
    \hline
    Unfiltered & BERT-base & $0.724 \pm 0.112$ \\
     & CLIP-BERT \\
     & \hspace{0.65cm}implicit & $0.744 \pm 0.080$ \\
     & \hspace{0.65cm}explicit & $0.870 \pm 0.086$ \\
     & \hspace{0.65cm}images & $0.876 \pm 0.063$ \\
     \hline
     Filtered & BERT-base & $0.460 \pm 0.083$ \\
     & CLIP-BERT \\
     & \hspace{0.65cm}implicit & $0.541 \pm 0.060$ \\
     & \hspace{0.65cm}explicit & $0.733 \pm 0.098$ \\
     & \hspace{0.65cm}images & $0.785 \pm 0.055$ \\
    \hline
    \end{tabular}
    \label{tab:memory-colors-results}
\end{table}

\paragraph{Are humans the top performers on the Memory Colors dataset?} 
We can conclude that the human baseline results are better than those of any model, training procedure and querying strategy evaluated. 
This baseline is expected to be high because the task is inherently based on the notions of color according to the majority of the humans that were evaluated. 

Furthermore, as language models lack much knowledge compared to humans, we expect them to perform worse than humans on this task.

Not even the CLIP-BERT model provided with gold standard images and unfiltered textual information on colors matches the performance of the human annotators.
There may be several reasons for this, for instance that the capacity of the multimodal models is not sufficient, or that humans are privy to additional information that helps them solve the Memory Colors task better.

\paragraph{Is the filtering of the training data necessary for our experimental setup of evaluating visual knowledge transfer?} 
We see that the BERT-base model without any further training has a bad performance on the Memory Colors dataset, only slightly better than the majority baseline.
On the other hand, the model shows significant performance improvement if it is trained on our unfiltered visual-and-language data. 
This suggests that the unfiltered training dataset contains much information about the objects' color \emph{textually}.
This is perhaps not surprising, as it is common that captions describe what colors the objects in the image have.
However, for our purposes it is problematic as we wish to constrain this information to be learnt from the visual modality solely.
Based on this, we conclude that the filtering of the training data is necessary in our experimental setup for evaluating visual knowledge transfer.

\paragraph{Does the filtering of the training data work as intended according to the Memory Colors dataset?} 
If we filter the training data of the BERT-base model, the performance drops from 0.724 to 0.460, indicating that a large portion of the necessary information has been removed.
However, the model performance does not drop to that of the original BERT-base model, so seemingly some color information still reaches the model through the text despite the data filtering. 
This leakage is undesirable from the perspective of evaluating visual knowledge transfer, since the model should not be able to perform well on Memory Colors without visual knowledge transfer capabilities.

\paragraph{Does the implicit transfer strategy improve performance on Memory Colors?}
The CLIP-BERT model using implicit transfer displays significantly better performance than the corresponding BERT-base baseline in the filtered case, while the performance difference is negligible in the unfiltered case.
This indicates that the implicit strategy does work to some extent, at least when corresponding textual information is lacking.

\paragraph{Does the explicit transfer strategy improve performance on Memory Colors?} 
The CLIP-BERT model using explicit transfer displays significantly better performance than the baseline and the implicit transfer model for both the unfiltered and filtered training methods. 
This suggests that the model has a strong visual knowledge transfer capability that enables it to improve the performance on Memory Colors, beyond the knowledge provided textually. 
However, we can observe that the rise from the baseline is larger for the filtered case than the unfiltered case, with a difference of 27 percentage units and 15 percentage units respectively. 
This is expected of our task, as the models need to rely more on their visual transfer capabilites to perform well in the filtered training case.

\paragraph{Is the explicit strategy better than the implicit?}
The fact that the performance is much improved over the text-only baseline also in the unfiltered case indicates that the explicit strategy indeed extends the textual knowledge in a complementary manner.
Since we do not see a similar performance gain for the implicit strategy, we have reason to believe that the explicit strategy is more effective.

\paragraph{What is the quality of the predicted visual features compared to those of the gold standard visuals?}
Lastly, we have the results of the CLIP-BERT model that bases its predictions on the ground truth image of each object it is being queried on. 
As expected, this model acts as an upper bound for the model performance on both the unfiltered and filtered method cases, while the rise in performance is more significant in the filtered training case. 
This also agrees with the previously mentioned hypothesis on the performance difference between the filtered and unfiltered training case. 
It also implies that the predicted features of the CLIP-BERT-explicit model are not as good as if they were generated from the actual item pictures.

\paragraph{Are the models sensitive to the phrasing of the query templates?}
The standard deviation figures presented in \Cref{tab:memory-colors-results} show the
variation in the accuracy scores for the different query templates.

We can observe that all of the models evaluated display a standard deviation between 5-11\%, and that none are lower than the standard deviation of the human baseline. Consequently, the models are sensitive to the phrasing of the query templates, 
as already mentioned in \Cref{sec:querying-strategies}.

\section{Related work}
There are multiple perspectives on how our contributions relate to previous work, and we elaborate on this in the subsequent sections.

\subsection{Visual grounding for improved NLP} 
A body of previous work exists on the topic of visual grounding for improving performance on language tasks.
For example, \citet{chrupala-etal-2015-learning} ground the language representations by adding an auxiliary visual feature prediction loss during training, and evaluate the learned representations on word and sentence similarity tasks.
Similarly, \citet{kiela-etal-2018-learning} align language and corresponding visual representations through a contrastive ranking loss, and evaluate the learned representations on a suite of common NLP classification tasks.
Visual grounding has also been explored for machine translation; for instance, \citet{elliott-kadar-2017-imagination} add an auxiliary visual prediction loss in addition to the regular seq2seq objective which is shown to improve performance.
More recently, \citet{sileo2021visual} investigates the extent to which visual-linguistic pretraining of multimodal transformers can improve performance on a set of text-only tasks.
While these approaches suggest that visual grounding can be helpful for language tasks, our work more explicitly targets the question of \emph{how} the additional modality can complement the textual signal.
We do this through a narrow focus on visual knowledge, in contrast to tasks requiring broader language understanding.

\subsection{Augmenting input using feature prediction}
Our work is not the first to implement the generation of imaginary features based on text for a unimodal text task. 
There is previous work investigating the potential of leveraging multimodal information during training to enable a model to generate or retrieve additional multimodal information at inference time for a pure text input.

\citet{sileo2021visual} uses the term \emph{associative grounding}, which can be based on synthesis or retrieval.
The main difference between our work and Sileo's is that he develops a model based on retrieval, while we use feature synthesis.
Earlier work has used 
latent visual features to augment the input for improving 
word embeddings~\citep{balanced_vqa_v2}.

The idea has been explored for non-visual information as well.
For example in open-domain question answering, retrieving relevant source documents as a preliminary step prior to knowledge extraction has proved highly effective~\citep{guu2020REALM}.
Recently, \citet{zellers2021piglet} also proposed a similar explicit decoupling but for augmenting a language model with knowledge about physical dynamics.
Also here, our work differs in that we augment the input with a visual signal and that we use it for a task focused on evaluating the capacity of visual knowledge transfer of a model.

\subsection{Visual-linguistic tasks} Much recent work on vision-and-text models focuses on developing models for multimodal tasks. 
Here, the model is queried with both textual and visual input on tasks such as VQA, GQA and NLVR2 ~\citep{balanced_vqa_v2,hudson2018gqa,suhr-etal-2019-corpus}. 
Recently developed models that can or could be found on the leaderboards of these tasks without using ensembling are e.g. ViLBERT, LXMERT, ImageBERT and OSCAR~\citep{NEURIPS2019_c74d97b0,tan-bansal-2019-lxmert,qi2020imagebert,li2020oscar}. 

These models are typically based on
the BERT transformer model architecture~\citep{devlin2018bert} and 
they often extract features from the visual input using a Faster R-CNN model~\citep{ren2015faster}. 
Similarly to this previous work, we also base our model design on the BERT model architecture and extract features from the visual input using a pre-trained visual processing model. 
However, we differ from the previous work in that we utilize the CLIP model to extract visual features, which also enables us to predict visual features from a pure textual input using the shared feature space for textual and visual representations of CLIP. 
We also differ in that we aim to study the visual knowledge transfer capabilities of a model by evaluating it with a method that measures visual knowledge for a unimodal textual task.

\section{Conclusions and future work}

We have introduced a methodology for measuring visual knowledge transfer in multimodal language models. The centerpiece is a new benchmark \emph{Memory Colors} designed to test how well such models incorporate knowledge about colors of common objects.

We find that careful filtering of the underlying training data can provide an effective means to attribute the acquired knowledge to the individual source modalities.

Our results based on this methodology also showcase that vision-and-language pre-trained language models are able to \emph{textually} express knowledge obtained from a separate (e.g. visual) modality.

We also found that there is some information leakage in our filtering method, as the performance of filtered BERT-base improves over the BERT-base baseline. 
This improvement in model performance cannot be explained based on the method and results of this work. Thus, it remains to be investigated what kind of 
information leakage takes place in spite of the filtering. 
Potential explanations
are that the model learns the color of an item 
through second-order effects, e.g. by learning the color of a synonymous item that we have not filtered for, or that the original BERT-base model already contains textual knowledge relevant to Memory Colors, while it needs further training on a visual-language dataset to access that knowledge.
Future work should ensure that the 
experimental setup for evaluating visual knowledge works as intended.

Additionally, it is worth investigating what other evaluation alternatives we have for measuring the cross-modal capabilities of NLP models. 
Can we create an evaluation methodology that is not task-based, can we find some other task to evaluate on, or can we improve on the statistical robustness of our evaluation methodology?

We observed that a model with implicit transfer performs better on our evaluation task than a unimodal language model, while a model with explicit transfer through prediction performs even better on the task. This implies that both implicit and explicit knowledge transfer are promising directions for efficient visual knowledge transfer to text, although the explicit transfer may be more promising.

While we here only investigate knowledge transfer from a visual modality, it is likely that this model design also can be successfully implemented for other modalities.

The experimental setup proposed in this work helped us discover and validate the potential of explicit transfer. We can conclude that more work on understanding how multimodal training of language models affect their predictions is an interesting direction towards more robust and trustworthy NLP systems.

\section*{Acknowledgments}

This work was partially supported by the Wallenberg  AI,  Autonomous  Systems  and  Software Program (WASP) funded  by  the  Knut  and  Alice Wallenberg Foundation.

Additionally, the computations were enabled by resources provided by the Swedish National Infrastructure for Computing (SNIC), partially funded by the Swedish Research Council through grant agreement no. 2018-05973.

Lastly, we would like to thank the 11 individuals who helped us annotate our Memory Colors dataset. Their work was imperative for the creation of this article. We also thank the anonymous reviewers for their valuable feedback and knowledge sharing.

\bibliographystyle{acl_natbib}
\bibliography{ref}

\pagebreak

\appendix

\section{Supplementary material}
\begin{figure}[h!]
    \centering
    \input{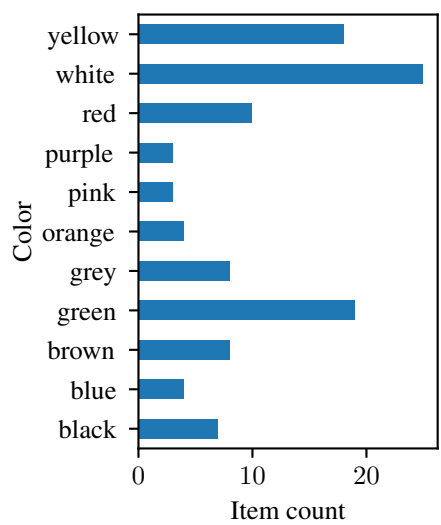}
    \caption{The color distribution of the 109 items in the Memory Colors dataset. The most frequent color in the dataset is \emph{white} with a count of 25. The colors with the lowest frequency are pink and purple, which only occur for 3 items each.}
    \label{fig:color-distribution}
\end{figure}

The supplementary material of this work includes the instructions provided to the human annotators in \Cref{fig:human-annotator-instructions}, the Memory Colors data in \Cref{tab:memory-colors-data}, the color label distribution of the data in \Cref{fig:color-distribution}, the different query templates for human annotators and models in \Cref{tab:query-templates} and the full statistics for the Vision-and-Language dataset in \Cref{tab:vlp_dataset_stats}. 

\vspace{0.5cm}
\begin{figure}[h!]
\begin{Verbatim}[frame=single, label=Annotator Instructions, fontsize=\small]
Thank you for helping us out by solving 
this task!

You will be presented with 121 
fill-in-the-gap color questions that 
are to be answered with one answer, 
where you can pick between the 
following answer alternatives:

yellow
blue
green
white
red
orange
black
pink
brown
grey
purple

You should fill in your answer for the 
gap [fill-in-this-word] under the 
column Fill-in-word. Answer with the 
alternative that first comes to your 
mind. The cell you are to fill in will 
turn green after you have specified one 
of the possible answers in it. Make 
sure that all cells in the column are 
green and not red before you submit 
your answers. Do not leave any cells 
empty, just guess on the alterative you 
find most likely even if you don't know 
the answer.

It is important that you solve this 
task by yourself, such that you do not 
discuss the questions or the answers 
with anyone else before you have 
submitted your answers. Also, you 
should not Google or look up anything 
while answering the questions.

Thank you again!

\end{Verbatim}
    \caption{The instructions provided to the human annotators before they annotated the predecessor to the Memory Colors dataset.}
    \label{fig:human-annotator-instructions}
\end{figure}

\begin{table*}[h!]
    \small
    \centering
    \caption{The 109 entries in the Memory Colors dataset.}
    \label{tab:memory-colors-data}
    \begin{subtable}[t]{0.45\textwidth}
    \begin{tabular}[t]{|c|l|l|l|}
    \hline
    Index & Descriptor & Item & Color \\
    \hline
    1 & a & sunflower & yellow  \\
2 & the & ocean & blue  \\
3 & & grass & green  \\
4 & & butter & yellow  \\
5 & & bone & white  \\
6 & & ivory & white  \\
7 & the & sky & blue  \\
8 & the inside of a & pineapple & yellow  \\
9 & a & tomato & red  \\
10 & a & strawberry & red  \\
11 & a & rose & red  \\
12 &  & blood & red  \\
13 & a & heart & red  \\
14 & a & pumpkin & orange  \\
15 & a & carrot & orange  \\
16 &  & cheese & yellow  \\
17 & the & sun & yellow  \\
18 & a & lemon & yellow  \\
19 &  & corn & yellow  \\
20 & a & frog & green  \\
21 & a & leaf & green  \\
22 & a & blueberry & blue  \\
23 &  & jeans & blue  \\
24 & the animal & bat & black  \\
25 & a & crow & black  \\
26 & a & raven & black  \\
27 &  & coal & black  \\
28 &  & paper & white  \\
29 &  & sugar & white  \\
30 &  & milk & white  \\
31 &  & snow & white  \\
32 &  & sheep & white  \\
33 & a & flamingo & pink  \\
34 &  & cherry blossoms & pink  \\
35 &  & soil & brown  \\
36 &  & stone & grey  \\
37 & an & elephant & grey  \\
38 & the animal & seal & grey  \\
39 & a & plum & purple  \\
40 &  & lavender & purple  \\
41 & a & polar bear & white  \\
42 & the inside of a & watermelon & red  \\
43 &  & honey & yellow  \\
44 & a & banana & yellow  \\
45 & an & orange & orange  \\
46 & a & pear & green  \\
47 & the fruit & mandarin & orange  \\
48 & a & cherry & red  \\
49 &  & salt & white  \\
50 & a & swan & white  \\
51 & a & snow leopard & white  \\
52 & an & arctic fox & white  \\
53 &  & steel & grey  \\
54 &  & clouds & white  \\
55 &  & rainclouds & grey  \\
\hline
    \end{tabular}
    \label{tab:memory-colors-data-l}
\end{subtable}
\begin{subtable}[t]{0.45\textwidth}
    \begin{tabular}[t]{|c|l|l|l|}
    \hline
    Index & Descriptor & Item & Color \\
    \hline
56 &  & plants & green  \\
57 & a & suit & black  \\
58 &  & cocoa & brown  \\
59 &  & chocolate & brown  \\
60 &  & concrete & grey  \\
61 &  & aluminium foil & grey  \\
62 & a & pea & green  \\
63 & a & rainforest & green  \\
64 &  & rice & white  \\
65 &  & pasta & yellow  \\
66 &  & spinach & green  \\
67 &  & broccoli & green  \\
68 & a & lime & green  \\
69 &  & guacamole & green  \\
70 &  & salmon meat & pink  \\
71 &  & yoghurt & white  \\
72 &  & cottage cheese & white  \\
73 &  & feta cheese & white  \\
74 &  & matcha & green  \\
75 &  & seaweed & green  \\
76 &  & garlic & white  \\
77 & an & aubergine & purple  \\
78 &  & ivy & green  \\
79 & a & ruby & red  \\
80 &  & flour & white  \\
81 &  & baking soda & white  \\
82 & a & snowman & white  \\
83 &  & gravel & grey  \\
84 & an & egg yolk & yellow  \\
85 & an & egg & white  \\
86 &  & moss & green  \\
87 &  & cinnamon & brown  \\
88 & the outside of a & coconut & brown  \\
89 &  & scrambled eggs & yellow  \\
90 & a & cucumber & green  \\
91 & a & fire extinguisher & red  \\
92 & a & duckling & yellow  \\
93 & a & panther & black  \\
94 & a & pine tree & green  \\
95 & a & tooth & white  \\
96 &  & feces & brown  \\
97 &  & urine & yellow  \\
98 & an & iceberg & white  \\
99 & a & school bus & yellow  \\
100 & a & chick & yellow  \\
101 &  & sails & white  \\
102 &  & wood & brown  \\
103 & a & lady bug & red  \\
104 & a & daffodil & yellow  \\
105 & a & dandelion & yellow  \\
106 &  & cardboard & brown  \\
107 & a & blackboard & black  \\
108 &  & basil & green  \\
109 &  & parsley & green  \\
\hline
    \end{tabular}
    \label{tab:memory-colors-data-r}
\end{subtable}
\end{table*}

\begin{table*}[htbp!]
    \centering
    \caption{The query templates used to query both human annotators and models on the Memory Colors task.}
    \label{tab:query-templates}
    \begin{subtable}[t]{0.75\textwidth}
    \caption{The question templates used to query the human annotators on the object-colors evaluation task.}
    \begin{tabular}{|c|l|}
    \hline
    Index & Template \\
    \hline
    1 & Q: What is the color of [DESCRIPTOR] [ITEM]? A: It is [MASK]. \\
    2 & What is the color of [DESCRIPTOR] [ITEM]? [MASK]. \\
    3 &The color of [DESCRIPTOR] [ITEM] is [MASK]. \\
    4 & The usual color of [DESCRIPTOR] [ITEM] is [MASK]. \\
    5 & {[}DESCRIPTOR] [ITEM] usually has the color of [MASK]. \\
    6 & What is the usual color of [DESCRIPTOR] [ITEM]? [MASK]. \\
    7 & What is the typical color of [DESCRIPTOR] [ITEM]? [MASK]. \\
    \hline
    \end{tabular}
    \label{tab:human-question-templates}
\end{subtable}\vspace{0.5cm}
\begin{subtable}[t]{0.8\textwidth}
\caption{The question templates used to query the models on the object-colors evaluation task.}
    \begin{tabular}{|c|l|}
    \hline
    Index & Template \\
    \hline
    1 & Q: What is the color of [DESCRIPTOR] [ITEM]? A: It is [MASK]. \\
    2 & Q: What is the color of [DESCRIPTOR] [ITEM]? [SEP] A: It is [MASK]. \\
    3 & Q: What is the colour of [DESCRIPTOR] [ITEM]? A: It is [MASK]. \\
    4 & What is the color of [DESCRIPTOR] [ITEM]? [MASK]. \\
    5 & What is the color of [DESCRIPTOR] [ITEM]? [SEP] [MASK]. \\
    6 & What is the colour of [DESCRIPTOR] [ITEM]? [MASK]. \\
    7 & The color of [DESCRIPTOR] [ITEM] is [MASK]. \\
    8 & The usual color of [DESCRIPTOR] [ITEM] is [MASK]. \\
    9 & {[}DESCRIPTOR] [ITEM] usually has the color of [MASK]. \\
    10 & What is the usual color of [DESCRIPTOR] [ITEM]? [MASK]. \\
    11 & What is the usual color of [DESCRIPTOR] [ITEM]? [SEP] [MASK]. \\
    12 & What is the typical color of [DESCRIPTOR] [ITEM]? [MASK]. \\
    13 & What is the typical color of [DESCRIPTOR] [ITEM]? [SEP] [MASK]. \\
    \hline
    \end{tabular}
    \label{tab:model-question-templates}
    \end{subtable}
\end{table*}

\begin{table*}[t]
\caption{\label{tab:vlp_dataset_stats} Total number of image and caption samples, in each respective source dataset. In Visual Genome QA, the ``caption'' is the concatenation of the question and answer strings. Since some image links in SBU Captions and Conceptual Captions have become broken, the total number of samples don't match what was originally reported. There are more captions than images in the dataset since several different captions may refer to the same image.}
\begin{tabular}{|l|llll|l|}
\hline
                        & \textbf{MS COCO} & \textbf{SBU Captions} & \textbf{VG QA} & \textbf{Conc. Captions} & \textbf{Total}      \\ \hhline{|=|====|=|}
\textbf{Validation}     &         &              &                  &                     &                     \\
\hspace{2mm} Captions   & 25,014  & 10,000       & 10,000           & 13,923              & 58,937              \\
\hspace{2mm} Images     & 5,000   & 10,000       & 9,311            & 13,923              & 38,234              \\

\textbf{Training-unfiltered}  &   &              &                  &                     &                     \\
\hspace{2mm} Captions   & 591,753 & 852,504      & 1,435,322        & 1,841,392           & 4,720,971           \\
\hspace{2mm} Images     & 118,287 & 852,504      & 99,255           & 1,841,392           & 2,911,438           \\

\textbf{Training-filtered} &         &              &                  &                     &                     \\
\hspace{2mm} Captions   & 548,887 & 760,724      & 1,348,120        & 1,771,940           & 4,429,671           \\
\hspace{2mm} Images     & 117,897 & 760,724      & 99,051           & 1,771,940           & 2,749,612           \\  \hline
\end{tabular}
\end{table*}

\end{document}